\documentclass{article}

\usepackage[preprint]{neurips_2026}

\usepackage[utf8]{inputenc} 
\usepackage[T1]{fontenc}    
\usepackage{hyperref}       
\usepackage{url}            
\usepackage{booktabs}       
\usepackage{amsfonts}       
\usepackage{nicefrac}       
\usepackage{microtype}      
\usepackage{xcolor}         
\usepackage{amsmath}
\usepackage{amssymb}
\usepackage{graphicx}
\usepackage[ruled,linesnumbered]{algorithm2e}
\usepackage{multirow}       
\usepackage{wrapfig}        
\usepackage{enumitem}       
\usepackage[most,minted]{tcolorbox}
\usepackage{minted}
\usepackage{xspace}

\definecolor{codebg}{HTML}{F8F8F8}
\definecolor{codeframe}{HTML}{BDBDBD}

\lstdefinelanguage{json}{
  basicstyle=\scriptsize\ttfamily,
  morestring=[b]",
  stringstyle=\color[HTML]{008000},
  literate=
    {:}{{{\color[HTML]{CC0000}{:}}}}{1}
    {,}{{{\color[HTML]{CC0000}{,}}}}{1}
    {\{}{{{\color[HTML]{0000CC}{\{}}}}{1}
    {\}}{{{\color[HTML]{0000CC}{\}}}}}{1}
    {[}{{{\color[HTML]{0000CC}{[}}}}{1}
    {]}{{{\color[HTML]{0000CC}{]}}}}{1},
}

\newtcblisting{jsonbox}[1][]{
  enhanced, breakable,
  colback=codebg, colframe=codeframe, boxrule=0.6pt, arc=2pt,
  left=2pt, right=2pt, top=4pt, bottom=4pt,
  fonttitle=\small\bfseries, coltitle=black, colbacktitle=codebg,
  attach boxed title to top left={xshift=6pt, yshift=-2mm},
  title={#1},
  listing engine=listings,
  listing options={
    language=json,
    basicstyle=\scriptsize\ttfamily,
    breaklines=true,
    breakatwhitespace=false,
    columns=flexible,
    keepspaces=true,
    showstringspaces=false,
  },
  listing only,
}

\newtcolorbox{codebox}[1][]{
  enhanced, breakable,
  colback=codebg, colframe=codeframe, boxrule=0.6pt, arc=2pt,
  left=6pt, right=6pt, top=4pt, bottom=4pt,
  fonttitle=\small\bfseries, coltitle=black, colbacktitle=codebg,
  attach boxed title to top left={xshift=6pt, yshift=-2mm},
  title=#1
}

\newcommand{\method}{Co-Coder\xspace}
\newcommand{\devbench}{DevEval\xspace}

\title{When Parallelism Pays Off: Cohesion-Aware\\ Task Partitioning for Multi-Agent Coding}

\author{
Xu Yang\textsuperscript{*}$^{1}$, Lunyiu Nie\textsuperscript{*}$^{1}$, Ethan Chandra$^{1}$, Stanislav Gannutin$^{1}$\\
\textbf{Fangru Lin}$^{2}$, \textbf{Swarat Chaudhuri}$^{1}$\\ \\
$^{1}$The University of Texas at Austin\ \ \ \  $^{2}$University of Oxford
}

\begin{document}

\maketitle

\begin{abstract}
  Multi-agent Large Language Model (LLM) systems offer a way to decompose complex tasks, such as coding, through parallelization and context isolation. However, adding agents in practice introduces inter-agent communication overhead, which incurs extra cost and can sometimes offset the efficiency gains.
  We formalize multi-agent orchestration as a graph partitioning problem that captures the \emph{communication-to-computation trade-off}: task decomposition can shorten critical-path computation, but cross-agent dependencies require costly context transfer.
  We instantiate this view in repository-level software engineering and present Cohesion-aware Coder (\method), which builds dependency graphs from static analysis, isolates structural hub files, partitions the graph via community detection, and executes the partition with a dependency-aware scheduler.
  Across $28$ real-world tasks on \devbench and CodeProjectEval, \method advances the Pareto-frontier over sequential and file-based parallel baselines as well as Claude Code with Agent Teams, lifting pass rate by up to $14.0\%$, achieving up to a $2.10\times$ wall-clock speedup, and reducing API cost by up to $35\%$, with the largest gains on the most dependency-dense projects.
  \method demonstrates how cohesion-aware orchestration can make parallel coding agents both theoretically grounded and practically efficient, suggesting a broader design principle for multi-agent systems.

  \let\thefootnote\relax\footnotetext{\textsuperscript{*} Equal Contribution. We release our code at \url{https://github.com/Flitternie/CoCoder}.}
\end{abstract}

\section{Introduction}
Large Language Model (LLM) agents are increasingly deployed not as monolithic solvers but as \emph{multi-agent systems} that decompose a task across several specialized workers running in parallel \cite{hong2024metagpt, qian2024chatdev, wu2023autogen, zhuge2024gptswarm, qian2025macnet}. The appeal of this paradigm is straightforward: more agents should mean more throughput, more structured context management, and shorter wall-clock latency. In practice, however, these benefits do not always hold: recent findings show that adding agents can degrade both speed and task performance compared to a single capable agent \cite{cemri2025why, malfa2025large, khatua2026cooperbench, mieczkowski2026language}.

We argue that this behavior is not an idiosyncrasy of LLMs but a familiar one, and that it is best understood through the lens of the \emph{communication-to-computation trade-off} that has organized four decades of distributed computing research \cite{valiant1990bridging, culler1993logp, graham1966bounds, kernighan1970efficient, karypis1998fast, topcuoglu2002performance}. In any distributed parallel system, total runtime is shaped by two competing terms: (i) the useful work each worker performs in isolation, and (ii) the coordination cost paid whenever one worker depends on another. Multi-agent systems are no exception: every cross-agent dependency must be resolved by transmitting context tokens, which inflate prompts, decoding time, and inference cost. While a growing body of work studies multi-agent communication topologies \cite{zhuge2024gptswarm, qian2025macnet} and parallel function calling \cite{kim2024llmcompiler}, the trade-off itself has not been formalized as an objective that an orchestrator can be evaluated against, and the structural problem of \emph{how tasks are partitioned across agents} in the first place has received little attention.

\begin{wrapfigure}{r}{0.46\linewidth}
  \vspace{-6pt}
  \centering
  \includegraphics[width=\linewidth]{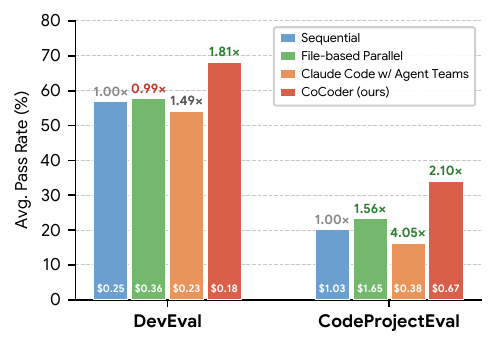}
  \caption{Average pass rate on \devbench and CodeProjectEval. Labels above bars: wall-clock speedup vs.\ Sequential version; labels inside bars: avg.\ API cost per task (USD).}
  \label{fig:intro_bar}
  \vspace{-8pt}
\end{wrapfigure}

We make this perspective concrete in the setting of repository-level software engineering, a canonical testbed in which task dependencies are unusually legible: files import from one another, call graphs are recoverable by static analysis, and the cost of cross-agent context transfer can be measured directly. The bottleneck of current multi-agent coding systems \cite{hong2024metagpt, qian2024chatdev, bairi2024codeplan} is precisely the trade-off above: when heavily interdependent files are assigned to different agents, those agents must repeatedly halt generation to query peers or ingest updated repository state, and wall-clock latency grows with the agent count rather than shrinking. The symptoms are visible in Figure~\ref{fig:intro_bar}: on CodeProjectEval, a file-based parallel system inflates API cost by 60\% for the $1.56\times$ efficiency gain; on two benchmarks, Claude Code with Agent Teams achieves the fastest wall-clock time but falls below the sequential version in terms of code quality. Both are failure modes of poorly-orchestrated parallelism: one wastes cost, the other sacrifices code quality.

Inspired by classic distributed computing systems, we propose \textbf{Co}hesion-aware \textbf{Coder} (\method), an orchestration framework that uses directed acyclic graphs to partition and schedule multi-agent systems. \method represents the project as a weighted dependency graph in which vertex weights encode per-file generation cost and edge weights encode the cost of transferring context across an agent boundary, and casts orchestration as a graph partitioning problem whose objective combines critical-path computation cost with cross-partition communication cost, the same two terms studied in classical distributed computing. Because code dependency graphs are directed, heterogeneous, and dominated by a few hub files (project-wide utilities and top-level entry points), generic spectral methods are a poor fit; \method instead isolates these hubs, then applies the Infomap algorithm \cite{rosvall2008maps}, which minimizes the description length of a random walk on the directed graph and thereby acts as a direct surrogate for cross-partition information flow. A final post-processing step recovers latent intra-cluster parallelism, and the resulting partition is executed by a dependency-aware list scheduler \cite{graham1966bounds} that exposes parallelism without global synchronization barriers.

Across $28$ real-world projects on two repository-level coding benchmarks, \method advances the Pareto frontier across pass rate, latency, and cost. On \devbench \cite{deveval}, it lifts average pass rate from 56.8\% to 68.1\% over the sequential version while delivering a $1.81\times$ speedup and a 28\% cost reduction; on the more challenging CodeProjectEval \cite{zhao2025realisticprojectlevelcodegeneration}, the gains widen to a 14.0\% increase in pass rate, a $2.10\times$ speedup, and a 35\% cost reduction. The gains are largest on projects with the densest cross-file dependencies, where baseline systems including Claude Code with Agent Teams either inflate cost without improving quality or sacrifice quality for speed, confirming that the communication-to-computation trade-off, rather than raw concurrency, is what governs multi-agent system behavior. 

In summary, our contributions are:
\begin{itemize}[leftmargin=1.5em,itemsep=2pt,topsep=2pt]
    \item We highlight the communication-to-computation trade-offs in multi-agent systems and formulate an objective that combines critical-path computation cost with inter-agent communication cost.
    \item We propose \method, an orchestrator that instantiates this objective via cohesion-based graph partitioning and dependency-aware scheduling on repository-level multi-agent coding tasks.
    \item We show empirically that \method advances the Pareto-frontier over sequential, file-parallel, and Claude Code baselines, with the largest gains on the most dependency-dense projects.
\end{itemize}

\begin{figure}[t]
\centering
\includegraphics[width=\textwidth]{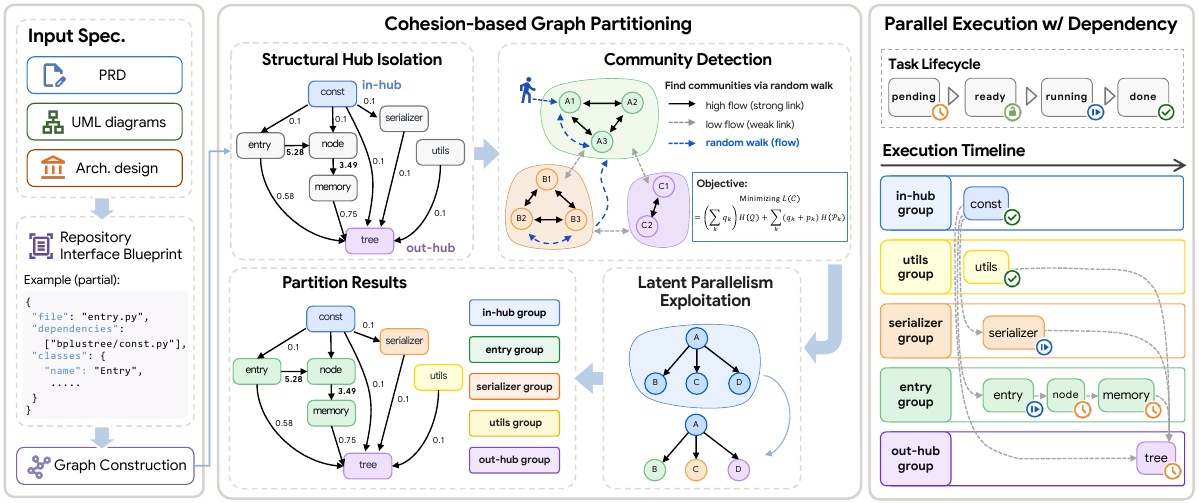}
\caption{Overview of \method. The input specifications from the benchmarks are first parsed into a Repository Interface Blueprint that encodes file-level dependencies, which subsequently constructs the weighted graph. We then perform cohesion-based partitioning over the graph via structural hub isolation, community detection, and latent parallelism exploitation. Finally, on the right side, parallel coding agents execute in a dependency-aware timeline, where multiple tasks across different groups follow a lifecycle and may fire in parallel once their upstream dependencies are completed. Note that dependency arrows are reversed for illustration: arrows point from dependency to dependent, opposite to the actual dependency graph convention.}
\label{fig:pipeline}
\end{figure}

\section{Problem Formulation}
\label{sec:formulation}

We formalize a type of systems in which $K$ parallel LLM agents jointly complete a composite task. The task decomposes into a set of subtasks with dependencies among them: some subtasks can only begin once others have been completed, and agents that work on dependent subtasks must share context about one another's outputs. The system's end-to-end overheads are therefore shaped by two competing factors: the time agents spend doing productive work in isolation, and the communication overhead they incur when resolving cross-agent dependencies. We formalize this tension by representing the task structure as a weighted dependency graph and casting orchestration as the problem of partitioning that graph to minimize a joint cost over both terms.

\paragraph{Task Structure as a Weighted Dependency Graph.}
We represent the composite task as a directed graph $G = (V, E)$. Each vertex $v_i \in V$ corresponds to one elementary subtask assigned to a single agent. A directed edge $(v_i, v_j) \in E$ indicates that $v_i$ depends on $v_j$: the agent handling $v_i$ must observe some output of $v_j$ before it can complete its own work. Each vertex carries a non-negative weight $w_i$ quantifying the \emph{computation cost} of completing $v_i$, measured as the expected wall-clock time an agent spends working on it in isolation. Each edge carries a non-negative coupling weight $c_{ij}$ quantifying the \emph{communication cost} incurred when $v_i$ and $v_j$ are handled by different agents: a heavier $c_{ij}$ means the two subtasks share more output interface and would require more context transfer to coordinate across an agent boundary. The specific instantiation of $V$, $E$, $w_i$, and $c_{ij}$ for repository-level coding is described in Section~\ref{sec:graph}.

\paragraph{Partitioning as Agent Assignment.}
Orchestrating $K$ agents corresponds to partitioning $V$ into $K$ disjoint groups $P = \{P_1, \dots, P_K\}$, where all subtasks in $P_k$ are handled by the same agent. The partition $P$ induces a scheduled execution DAG $G_P = (V,\, E \cup E_P)$, where $E_P$ contains the intra-partition serialization edges that arise because each agent processes its subtasks one at a time. The system-wide \emph{computation cost} is the critical-path latency of this DAG,
\begin{equation}
\label{eq:Wp}
W(P) \;=\; \max_{\pi\,\in\,\Pi(G_P)}\, \sum_{v_i\, \in\, \pi}\, w_i,
\end{equation}
where $\Pi(G_P)$ is the set of directed paths in $G_P$. $W(P)$ captures two sources of serialization: the sequential workload within each agent's group, and cross-partition dependency chains that force a downstream agent to wait for an upstream subtask owned by a different agent. The system-wide \emph{communication cost} is the total coupling weight across partition boundaries,
\begin{equation}
\label{eq:Cp}
C(P) \;=\; \sum_{\substack{(v_i, v_j) \in E \\ k(v_i)\, \ne\, k(v_j)}} c_{ij},
\end{equation}
where $k(v_i)$ denotes the partition index of $v_i$. Coarser partitions internalize dependencies and reduce $C(P)$ but lengthen intra-agent chains and inflate $W(P)$; finer partitions expose parallelism and reduce $W(P)$ but push more dependencies across agent boundaries and inflate $C(P)$.

\paragraph{Optimization Objective.}
We seek the partition that minimizes the joint cost of both terms,
\begin{equation}
\label{eq:objective}
T(P) = W(P) + \alpha \cdot C(P), \qquad P^* = \arg\min_{P} T(P),
\end{equation}
where $\alpha > 0$ represents a trade-off weighting factor and $K = |P|$ is determined by the partition itself. This objective is the direct analogue of the communication-to-computation cost studied in classical distributed task scheduling~\cite{topcuoglu2002performance, valiant1990bridging}: $W(P)$ plays the role of the critical-path computation time, and $\alpha C(P)$ plays the role of inter-processor communication overhead. Section~\ref{sec:method} introduces \method as a principled approach to minimizing $T(P)$ in the setting of repository-level software engineering.

\section{The \method Framework}
\label{sec:method}

The objective $T(P)$ in Section~\ref{sec:formulation} is defined over a weighted dependency graph that is not given directly --- it must be inferred from the project's requirements before any partitioning can take place. Once the graph is available, the partition must balance two competing demands: grouping tightly coupled files to minimize inter-agent communication, and avoiding serialization bottlenecks to keep wall-clock time short. Finally, the partition must be realized by a scheduler that respects cross-partition dependencies without introducing unnecessary synchronization barriers. \method addresses each concern in sequence through three stages: \emph{graph construction} (Section~\ref{sec:graph}), \emph{cohesion-based partitioning} (Section~\ref{sec:partition}), and \emph{dependency-aware execution} (Section~\ref{sec:execution}).

\subsection{Graph Construction}
\label{sec:graph}

\paragraph{Repository Interface Blueprint.}
The graph $G$ is not directly observable from the input, which consists of free-form requirement documents ( product specifications, UML class and sequence diagrams). To bridge this gap, we prompt an LLM to produce a \emph{Repository Interface Blueprint} (RIB), a structured outline of the project at file granularity. For each file $f_i$, the RIB enumerates the symbols it defines (class names, function signatures with typed parameters, and global constants) and the set of files it imports from. The symbol inventory seeds the per-vertex computation weight $w_i$ from interface size, which serves as a proxy for generation latency. The import graph yields the edge set $E$. The RIB is generated by a single LLM call and refined through a lightweight self-critique loop (Appendix~\ref{sec:rib_validation}).

\paragraph{Edge Weights from Symbol Sharing.}
The RIB tells us which files depend on which, but not how strongly. We estimate the coupling weight $c_{ij}$ from the symbols the two files share: it should be large when one file defines symbols the other must consume, and small when both files merely co-reference symbols defined elsewhere. We capture this asymmetry by weighting each symbol higher in a file's feature vector when the file defines it than when it only references it, after excluding builtins and boilerplate (e.g., common library imports, pure type aliases, and module-level constants). For every edge $(v_i, v_j) \in E$ we set
\begin{equation}
\label{eq:cosweight}
c_{ij} = \gamma \cdot \frac{\mathbf{s}_i \cdot \mathbf{s}_j}{\|\mathbf{s}_i\|\, \|\mathbf{s}_j\|},
\end{equation}
where $\mathbf{s}_i$ is the sparse symbol vector for $v_i$ and $\gamma > 0$ calibrates the weights into the same units as $\alpha$ in Eq.~\eqref{eq:objective}. Definition–use edges receive a high$\times$low inner product and thus a large $c_{ij}$; edges of incidental co-occurrence receive a low$\times$low product and a small $c_{ij}$.

\subsection{Cohesion-based Graph Partitioning}
\label{sec:partition}

With the weighted dependency graph $G$ in hand, the next stage is to partition $G$ in a way that jointly addresses the two cost terms in $T(P) = W(P) + \alpha\, C(P)$ (Section~\ref{sec:formulation}): the critical-path latency $W(P)$ taken over the scheduled DAG $G_P = (V,\, E \cup E_P)$ that augments $E$ with intra-agent serialization edges $E_P$, and the cross-partition communication cost $C(P) = \sum_{(v_i,v_j)\in E,\,k(v_i)\ne k(v_j)} c_{ij}$. Both terms demand \emph{cohesion}: the degree to which files within a group are more tightly coupled to each other than to files outside it. High cohesion keeps the costliest dependencies inside group boundaries (suppressing inter-agent communication) while limiting the length of intra-group serialization chains (keeping execution on the critical path short). \method produces high-cohesion partitions through a three-step pipeline summarized in Algorithm~\ref{alg:partition}: (i) isolate structural hubs that obscure modular structure (lines 1--2); (ii) cluster the remaining graph into cohesive communities (lines 3--4); (iii) lift independent files out of their clusters to expose latent parallelism (lines 5--8).

\paragraph{Isolating Structural Hubs.}
Real code dependency graphs are highly heterogeneous: a small number of files act as project-wide utilities or top-level entry points \cite{martin2003agile}, and these have a disproportionate effect on $T(P)$. Vertices $v_i$ with high in-degree are widely depended-upon utilities (constants, shared schemas, exception hierarchies), which we call \emph{in-hubs}: their dependents are spread across most partitions, so the cut edges incident to an in-hub are nearly invariant to its assignment, while merging it into any group $P_k$ inserts it into that agent's serialization chain and inflates $W(P_k)$ by $w_i$. Singleton-isolating each in-hub therefore leaves $C(P)$ essentially unchanged but lifts $v_i$ off every group's critical path. Vertices with high out-degree are top-level aggregators and controllers, which we call \emph{out-hubs}: placed in any group $P_k$, an out-hub $v_i$ forces all of its in-group upstreams into a chain terminating at $v_i$, inflating $W(P_k)$ in proportion to the number of upstreams pulled in. Grouping all out-hubs into a single integration group concentrates this late-stage serialization at one agent without polluting the critical paths of the cohesive groups. Both hub types also bias generic clustering by collapsing the graph into one large community; isolating them first allows the remaining vertices to be clustered cleanly. The subgraph $G_C$ induced by the remaining non-hub vertices is passed to the next step.

\paragraph{Community Detection.}
With hubs removed, $G_C$ has a more uniform structure amenable to community detection. We cluster $G_C$ into cohesive groups using Infomap~\cite{rosvall2008maps}, a community-detection algorithm that operates natively on directed weighted graphs and infers the number of clusters from the data. Let $E_C$ be the edge set of $G_C$. Infomap models $G_C$ as a random walk whose transition probabilities follow the coupling-weighted topology:
\begin{equation}
\label{eq:transprob}
p(v_j \mid v_i) \;=\; \frac{c_{ij}}{\sum_{(v_i,\, v_l)\,\in\, E_C} c_{il}},
\end{equation}
so the walk tends to remain within strongly coupled modules and infrequently traverses low-weight cross-module edges. For a candidate partition $\mathcal{C} = \{C_1, \dots, C_m\}$, let $q_k = \sum_{v_i \in C_k,\, v_j \notin C_k} \pi_i\, p(v_j \mid v_i)$ be the exit rate of cluster $C_k$ and $p_k = \sum_{v_i \in C_k} \pi_i$ its stationary residence probability, where $\pi_i$ is the stationary visit probability of $v_i$. Infomap selects $\mathcal{C}$ by minimizing the two-level description length,
\begin{equation}
\label{eq:mapeq}
L(\mathcal{C}) \;=\; \Big(\sum_{k} q_k\Big) H(\mathcal{Q}) \;+\; \sum_{k} (q_k + p_k)\, H(\mathcal{P}_k),
\end{equation}
where $H(\mathcal{Q})$ is the entropy of the inter-cluster exit distribution and $H(\mathcal{P}_k)$ is the entropy of module $C_k$'s codebook over its internal node visits and the exit event. Intuitively, a high $H(\mathcal{Q})$ indicates frequent cross-cluster transitions (low cohesion), while a high $H(\mathcal{P}_k)$ indicates that an agent must process many weakly related files. Minimizing $L(\mathcal{C})$ therefore steers the partition away from cutting heavily weighted edges, effectively suppressing $C(P)$. The granularity of the partition is controlled by the Markov time $t$: at larger $t$ the random walk traverses more edges per step, making it easier to escape small modules and raising their $q_k$, so Infomap merges them into fewer, larger clusters. This reduces communication cost $C(P)$ at the expense of longer intra-group serialization chains (higher $W(P)$), providing a principled knob for navigating the computation--communication trade-off in Eq.~\eqref{eq:objective}.

\paragraph{Latent Parallelism Exploitation.}
Infomap produces cohesive clusters that suppress $C(P)$, but cohesion alone does not address $W(P)$: within a single cluster, vertices that share an upstream dependency but have no dependency on each other are still processed serially, unnecessarily stretching the critical path. We recover this latent parallelism by lifting any vertex $v_i$ that (i) has no intra-cluster dependents and (ii) shares its sole intra-cluster upstream $v_j$ with at least one cluster sibling. Lifting $v_i$ to a singleton converts the edge $(v_i, v_j)$ into a cut at cost $\alpha\, c_{ij}$, but removes $w_i$ from the cluster's serialization chain and shortens the critical path of $G_P$ by the same amount; condition (ii) ensures $v_j$ retains an intra-cluster dependent, so no other cluster edge is disturbed. Lifting therefore strictly reduces $T(P)$ whenever $\alpha\, c_{ij} < w_i$, which holds for liftable leaf vertices whose coupling to their upstream is outweighed by their generation cost $w_i$.

\begin{algorithm}[t]
\caption{Cohesion-based Graph Partitioning}
\label{alg:partition}
\DontPrintSemicolon
\KwIn{Weighted directed graph $G = (V, E, \{w_i\}, \{c_{ij}\})$ constructed from Section~\ref{sec:graph}}
\KwOut{Partition $P = \{P_1, \dots, P_K\}$ of $V$, one group per agent}
\BlankLine
$V_{\mathrm{in}} \leftarrow$ high in-degree vertices (\textit{in-hubs});\quad $V_{\mathrm{out}} \leftarrow$ high out-degree vertices (\textit{out-hubs});\;
$P \leftarrow \{\,\{v\} : v \in V_{\mathrm{in}}\,\} \cup \{V_{\mathrm{out}}\}$\tcp*{isolate hubs off critical paths}
$G_C \leftarrow G[\,V \setminus (V_{\mathrm{in}} \cup V_{\mathrm{out}})\,]$;\quad $\{C_1, \dots, C_m\} \leftarrow \mathrm{Infomap}(G_C)$\tcp*{partition by Eq.~\eqref{eq:mapeq}}
$P \leftarrow P \cup \{C_1, \dots, C_m\}$\;
\ForEach{cluster $C_k$}{
  $S_k \leftarrow \{\,v_i \in C_k :$ (i) no vertex in $C_k$ depends on $v_i$; (ii) $v_i$'s sole intra-cluster upstream $v_j \in C_k$ has another dependent in $C_k\,\}$\tcp*{leaves liftable to singletons}
  $C_k \leftarrow C_k \setminus S_k$;\quad $P \leftarrow P \cup \{\{v_i\} : v_i \in S_k\}$\tcp*{lift to exploit parallelism}
}
\Return{$P$}\;
\end{algorithm}

\subsection{Dependency-Aware Parallel Execution}
\label{sec:execution}

The partition $P$ assigns files to agents, but cross-partition edges still impose ordering constraints: an agent cannot faithfully generate a downstream file before its upstream dependencies are available. Naively gating execution on the topological depth of the dependency DAG forces every agent to wait at the slowest layer boundary and erodes the parallelism that the partition was designed to expose. \method instead executes the partition through a \emph{shared task list}, in which each file is registered as a task whose state advances from \emph{pending} to \emph{ready} the moment all of its upstream files have completed, and is then picked up by the agent that owns its group. There are no global synchronization barriers between dependency layers, and faster agents immediately move on to the next ready task. This realizes the standard greedy list-scheduling policy~\cite{graham1966bounds} on the partitioned graph and approaches the critical-path latency $W(P)$ defined in Eq.~\eqref{eq:Wp}.

After all files are generated, a leader agent runs the project's test suite and uses the partition record to localize each test failure to the group that owns the responsible file. Because partitions correspond to cohesive units of code, repair requests can be dispatched partition-by-partition rather than broadcast to the whole system: each agent reasons over a bounded slice of context, and conflicting edits across groups are avoided by construction. The repair loop iterates for a fixed number of rounds, with each round re-running the test suite and re-dispatching only the groups whose files still have failing tests.

\section{Experimental Setups}
\label{sec:exp_setup}

\subsection{Benchmarks}
\label{sec:benchmarks}

We evaluate on two repository-level code generation benchmarks. \textbf{\devbench}~\cite{deveval} consists of compact, self-contained repositories for end-to-end agent pipelines; we use its Python subset of $10$ projects, whose ground-truth reference implementations average $3.1$ files and $243$ LOC. \textbf{CodeProjectEval}~\cite{zhao2025realisticprojectlevelcodegeneration} is curated from real-world open-source libraries and contributes $18$ projects, with reference implementations averaging $11.9$ files and $2{,}371$ LOC per project, targeting deep cross-file dependencies and large integration surfaces. Both benchmarks provide unit tests for automated evaluation.

\subsection{Metrics}
\label{sec:metrics}

We report three per-repository metrics: (i)~\emph{pass rate}, the fraction of held-out unit tests passed; (ii)~\emph{wall-clock latency} in seconds, measured from pipeline start to the completion of the final agent, excluding testing; and (iii)~\emph{API cost} in USD, summed across all involved agents' input and output token usage. Each repository is evaluated over three independent runs and we report mean values.

For aggregate comparison, we normalize latency and cost against the sequential version: we compute the ratio of a method's per-task average latency/cost across all repositories to that of the sequential baseline's mean on the same benchmark. Both ratios are lower-is-better.

\subsection{Baselines}
\label{sec:baselines}

We compare our system against the following methods, all of which receive the same input specifications, parsed into a shared Repository Interface Blueprint: (a)~\textbf{Sequential}, where a single coding agent generates the entire repository,  autonomously deciding task order and managing its own context; (b)~\textbf{File-based Parallel}, where, from the same blueprint, a leader agent spawns one coding agent per file and the agents run concurrently and may communicate with each other, but no structural partitioning or dependency-aware scheduling is applied; (c)~\textbf{Claude Code w/ Agent Teams}, where, as an external reference, we run Anthropic's Claude Code CLI in its native agent-team mode, in which a team lead spawns and coordinates independent Claude Code sessions that message each other directly; and (d)~\textbf{\method (ours)}, our full system (Section~\ref{sec:method}) with cohesion-based graph-partitioning over the same blueprint to guide the multi-agent orchestration.

For a fair comparison, all methods use \texttt{gpt-5-mini} as the base model. Sequential, File-based Parallel, and \method are built on a consistent OpenHands SDK $\texttt{v1.11.4}$. Claude Code runs at $\texttt{v2.1.119}$ with LiteLLM $\texttt{v1.83.8}$ as the API proxy.

\section{Results}
\label{sec:results}

\newcommand{\better}[1]{\textcolor[HTML]{1B7F3A}{#1}}
\newcommand{\worse}[1]{\textcolor[HTML]{B00020}{#1}}

\begin{table}[t]
\centering
\caption{Per-project results on \devbench. \textit{\#LOC} represents lines of code in the ground truth implementation; \textit{\#Tests} is the total number of unit tests; \textit{\#Pass} reports the number of unit tests passed; \textit{Time} represents wall-clock latency in seconds; \textit{Cost} is total token cost in USD. Values denote mean$_{\pm\text{sd}}$ over three runs. Best \#Pass per row in \textbf{bold}; The Avg row reports per-task mean pass rate (\%); the $\times$ values are per-task means of Time and Cost normalized to Sequential (lower is better), with \better{green} indicating better than Sequential and \worse{red} indicating worse.}
\label{tab:deveval_perrepo}
\setlength{\tabcolsep}{3pt}
\footnotesize
\resizebox{\textwidth}{!}{%
\begin{tabular}{l c c ccc ccc ccc ccc}
\toprule
\multirow{2}{*}{Task} & \multirow{2}{*}{\#LOC} & \multirow{2}{*}{\#Tests} & \multicolumn{3}{c}{Sequential} & \multicolumn{3}{c}{File-based Parallel} & \multicolumn{3}{c}{Claude Code w/ Agent Teams} & \multicolumn{3}{c}{\textbf{\method (ours)}} \\
\cmidrule(lr){4-6}\cmidrule(lr){7-9}\cmidrule(lr){10-12}\cmidrule(lr){13-15}
 & & & \#Pass & Time & Cost & \#Pass & Time & Cost & \#Pass & Time & Cost & \#Pass & Time & Cost \\
\midrule
ArXiv\_digest    & 198 & 38 & 30.0$_{\pm 0.0}$ & 449$_{\pm 44}$ & 0.13$_{\pm 0.02}$ & 29.7$_{\pm 0.5}$ & \worse{782}$_{\pm 116}$ & \worse{0.23}$_{\pm 0.06}$ & 20.0$_{\pm 14.1}$ & \worse{573}$_{\pm 560}$ & \worse{0.21}$_{\pm 0.17}$ & \textbf{31.3$_{\pm 0.5}$} & \worse{475}$_{\pm 79}$ & \better{0.11}$_{\pm 0.01}$ \\
chakin           & 62 & 1 & \textbf{1.0$_{\pm 0.0}$} & 394$_{\pm 45}$ & 0.15$_{\pm 0.02}$ & \textbf{1.0$_{\pm 0.0}$} & \worse{456}$_{\pm 220}$ & \worse{0.16}$_{\pm 0.01}$ & 0.7$_{\pm 0.5}$ & \better{150}$_{\pm 84}$ & \better{0.09}$_{\pm 0.04}$ & \textbf{1.0$_{\pm 0.0}$} & \better{319}$_{\pm 56}$ & \better{0.11}$_{\pm 0.02}$ \\
geotext          & 134 & 4 & 2.0$_{\pm 0.8}$ & 740$_{\pm 213}$ & 0.37$_{\pm 0.21}$ & 2.0$_{\pm 0.0}$ & \worse{779}$_{\pm 130}$ & \better{0.30}$_{\pm 0.06}$ & 1.7$_{\pm 1.2}$ & \better{587}$_{\pm 436}$ & \worse{0.48}$_{\pm 0.42}$ & \textbf{2.7$_{\pm 0.5}$} & \better{691}$_{\pm 290}$ & \better{0.29}$_{\pm 0.14}$ \\
hone             & 274 & 7 & 2.3$_{\pm 1.2}$ & 994$_{\pm 83}$ & 0.37$_{\pm 0.08}$ & 2.0$_{\pm 0.0}$ & \better{556}$_{\pm 52}$ & \worse{0.43}$_{\pm 0.08}$ & 4.0$_{\pm 1.6}$ & \better{218}$_{\pm 95}$ & \better{0.15}$_{\pm 0.06}$ & \textbf{6.0$_{\pm 0.0}$} & \better{441}$_{\pm 32}$ & \better{0.24}$_{\pm 0.02}$ \\
Hybrid\_Images   & 144 & 19 & \textbf{17.0$_{\pm 0.0}$} & 296$_{\pm 56}$ & 0.05$_{\pm 0.01}$ & \textbf{17.0$_{\pm 0.0}$} & \better{210}$_{\pm 50}$ & \worse{0.07}$_{\pm 0.00}$ & \textbf{17.0$_{\pm 0.0}$} & \worse{798}$_{\pm 51}$ & \worse{0.25}$_{\pm 0.06}$ & \textbf{17.0$_{\pm 0.0}$} & \better{179}$_{\pm 45}$ & 0.05$_{\pm 0.01}$ \\
lice             & 376 & 25 & 8.0$_{\pm 0.0}$ & 514$_{\pm 87}$ & 0.16$_{\pm 0.02}$ & 9.0$_{\pm 1.4}$ & \worse{520}$_{\pm 101}$ & \worse{0.18}$_{\pm 0.05}$ & 8.7$_{\pm 0.9}$ & \better{138}$_{\pm 52}$ & \better{0.10}$_{\pm 0.05}$ & \textbf{9.7$_{\pm 1.7}$} & \better{325}$_{\pm 95}$ & \better{0.10}$_{\pm 0.03}$ \\
pso              & 168 & 5 & \textbf{5.0$_{\pm 0.0}$} & 472$_{\pm 85}$ & 0.13$_{\pm 0.01}$ & \textbf{5.0$_{\pm 0.0}$} & \better{345}$_{\pm 47}$ & \worse{0.14}$_{\pm 0.01}$ & \textbf{5.0$_{\pm 0.0}$} & \worse{581}$_{\pm 333}$ & \worse{0.26}$_{\pm 0.12}$ & \textbf{5.0$_{\pm 0.0}$} & \better{202}$_{\pm 11}$ & \better{0.08}$_{\pm 0.01}$ \\
readtime         & 284 & 8 & \textbf{2.0$_{\pm 0.0}$} & 853$_{\pm 133}$ & 0.23$_{\pm 0.02}$ & \textbf{2.0$_{\pm 0.0}$} & \better{486}$_{\pm 168}$ & \worse{0.25}$_{\pm 0.03}$ & 1.3$_{\pm 0.9}$ & \better{643}$_{\pm 366}$ & \better{0.13}$_{\pm 0.10}$ & \textbf{2.0$_{\pm 0.0}$} & \better{400}$_{\pm 101}$ & \better{0.14}$_{\pm 0.01}$ \\
stocktrends      & 384 & 7 & 3.0$_{\pm 0.0}$ & 2257$_{\pm 304}$ & 0.70$_{\pm 0.05}$ & 3.0$_{\pm 0.0}$ & \worse{3412}$_{\pm 372}$ & \worse{1.55}$_{\pm 0.12}$ & \textbf{4.3$_{\pm 1.9}$} & \better{1251}$_{\pm 232}$ & \better{0.52}$_{\pm 0.11}$ & 3.0$_{\pm 0.0}$ & \better{885}$_{\pm 68}$ & \better{0.50}$_{\pm 0.02}$ \\
TextCNN          & 403 & 10 & 1.7$_{\pm 1.2}$ & 1029$_{\pm 111}$ & 0.25$_{\pm 0.05}$ & 2.7$_{\pm 2.0}$ & \better{518}$_{\pm 187}$ & \worse{0.32}$_{\pm 0.02}$ & 2.0$_{\pm 1.4}$ & \better{424}$_{\pm 279}$ & \better{0.15}$_{\pm 0.09}$ & \textbf{5.0$_{\pm 0.0}$} & \better{504}$_{\pm 148}$ & \better{0.21}$_{\pm 0.02}$ \\
\midrule
Avg              & -- & -- & 56.8\% & 800 ($1\times$) & 0.25 ($1\times$) & 57.7\% & 806 (\worse{$1.01\times$}) & 0.36 (\worse{$1.44\times$}) & 54.1\% & 536 (\better{$0.67\times$}) & 0.23 (\better{$0.92\times$}) & \textbf{68.1\%} & \textbf{442 (\better{$0.55\times$})} & \textbf{0.18 (\better{$0.72\times$})} \\
\bottomrule
\end{tabular}}
\end{table}

\paragraph{\devbench.} As shown in Table~\ref{tab:deveval_perrepo}, \method achieves the highest average pass rate (68.1\%) among all methods, improving over the sequential baseline by 11.3\% while reducing wall-clock latency by 45\% (442\,s vs.\ 800\,s) and API cost by 28\% (\$0.18 vs.\ \$0.25). On simpler projects where a single agent already succeeds (e.g., \texttt{Hybrid\_Images}, \texttt{pso}), all four methods perform comparably. The advantage of \method emerges on projects requiring cross-file coordination: on \texttt{hone}, \method passes 6.0/7 tests compared to 2.3 for the sequential baseline, and on \texttt{TextCNN} it reaches 5.0/10 versus 1.7. This suggests that cohesion-aware task partitioning helps agents produce more consistent implementations across tightly coupled files.

File-based Parallel fails in reducing latency compared to the Sequential version on \devbench (806\,s vs.\ 800\,s) yet inflates cost by 44\% (\$0.36 vs.\ \$0.25) for a negligible pass-rate gain (57.7\% vs.\ 56.8\%). The poor task partitioning causes parallel agents to produce conflicting interfaces, requiring redundant re-generation that incurs extra costs without improving code quality. Claude Code w/ Agent Teams reduces latency (536\,s) but suffers the lowest pass rate (54.1\%), indicating that self-coordinated inter-agent orchestration cannot substitute for \method's explicit cohesion-aware task partitioning.

\paragraph{CodeProjectEval.} On the more challenging CodeProjectEval benchmark (Table~\ref{tab:cpe_perrepo}), the performance gap widens considerably. \method achieves 34.1\% average pass rate, a gain of 14.0\% over the sequential baseline (20.1\%) and 10.8\% over File-based Parallel (23.3\%). Meanwhile, \method reduces latency by 52\% (1315\,s vs.\ 2756\,s) and cost by 35\% (\$0.67 vs.\ \$1.03) relative to the sequential baseline. These gains are most pronounced on projects with deep dependency structures: on \texttt{simpy}, \method passes 58.0/149 tests while Sequential passes only 10.3; on \texttt{pyjwt}, \method reaches 140.0/294 versus 79.0.

The failure mode of na\"ive parallelism is particularly evident on this benchmark. File-based Parallel increases cost by 60\% (\$1.65 vs.\ \$1.03) relative to Sequential yet gains only 3.2\% in pass rate, since concurrently generated files frequently violate cross-file type contracts, leading to expensive yet unproductive iterations. Claude Code w/ Agent Teams achieves the lowest latency (680\,s) and cost (\$0.38) but again at the expense of the lowest pass rate (16.3\%), falling below even the sequential baseline. This confirms that aggressive parallelism without structural guidance degrades code quality on complex projects where inter-agent dependencies are critical.

\begin{table}[t]
\centering
\caption{Per-project results on CodeProjectEval. Metrics and conventions are the same as in Table~\ref{tab:deveval_perrepo}.}
\label{tab:cpe_perrepo}
\setlength{\tabcolsep}{3pt}
\footnotesize
\resizebox{\textwidth}{!}{%
\begin{tabular}{l c c ccc ccc ccc ccc}
\toprule
\multirow{2}{*}{Task} & \multirow{2}{*}{\#LOC} & \multirow{2}{*}{\#Tests} & \multicolumn{3}{c}{Sequential} & \multicolumn{3}{c}{File-based Parallel} & \multicolumn{3}{c}{Claude Code w/ Agent Teams} & \multicolumn{3}{c}{\textbf{\method (ours)}} \\
\cmidrule(lr){4-6}\cmidrule(lr){7-9}\cmidrule(lr){10-12}\cmidrule(lr){13-15}
 & & & \#Pass & Time & Cost & \#Pass & Time & Cost & \#Pass & Time & Cost & \#Pass & Time & Cost \\
\midrule
bplustree                       & 1{,}509 & 356 & 18.0$_{\pm 0.8}$ & 2192$_{\pm 236}$ & 0.99$_{\pm 0.11}$ & 22.0$_{\pm 9.3}$ & \worse{2312}$_{\pm 266}$ & \worse{1.30}$_{\pm 0.11}$ & 55.7$_{\pm 73.8}$ & \better{1055}$_{\pm 571}$ & \better{0.67}$_{\pm 0.55}$ & \textbf{73.0$_{\pm 53.7}$} & \better{1311}$_{\pm 189}$ & \better{0.56}$_{\pm 0.12}$ \\
cookiecutter                    & 2{,}805 & 375 & 59.0$_{\pm 6.7}$ & 4260$_{\pm 341}$ & 1.80$_{\pm 0.14}$ & 46.7$_{\pm 7.9}$ & \better{2180}$_{\pm 877}$ & \worse{2.37}$_{\pm 0.18}$ & 15.7$_{\pm 18.1}$ & \better{338}$_{\pm 246}$ & \better{0.21}$_{\pm 0.25}$ & \textbf{74.7$_{\pm 1.2}$} & \better{1693}$_{\pm 314}$ & \better{0.87}$_{\pm 0.11}$ \\
csvs-to-sqlite                  & 816 & 25 & 5.0$_{\pm 7.1}$ & 1525$_{\pm 183}$ & 0.46$_{\pm 0.05}$ & 10.7$_{\pm 7.5}$ & \worse{1739}$_{\pm 265}$ & \worse{0.68}$_{\pm 0.13}$ & \textbf{16.7$_{\pm 5.9}$} & \worse{1746}$_{\pm 1184}$ & \worse{0.51}$_{\pm 0.37}$ & 15.7$_{\pm 0.5}$ & \better{1020}$_{\pm 195}$ & \better{0.25}$_{\pm 0.02}$ \\
deprecated                      & 597 & 176 & 110.0$_{\pm 1.4}$ & 639$_{\pm 41}$ & 0.22$_{\pm 0.03}$ & 112.3$_{\pm 6.6}$ & \worse{1225}$_{\pm 275}$ & \worse{0.42}$_{\pm 0.06}$ & 76.7$_{\pm 54.2}$ & \better{140}$_{\pm 64}$ & \better{0.11}$_{\pm 0.08}$ & \textbf{115.7$_{\pm 5.2}$} & \worse{752}$_{\pm 63}$ & \better{0.21}$_{\pm 0.03}$ \\
drf-simplejwt   & 1{,}712 & 191 & 6.7$_{\pm 2.4}$ & 3633$_{\pm 200}$ & 1.42$_{\pm 0.11}$ & 4.7$_{\pm 0.5}$ & \better{2022}$_{\pm 493}$ & \worse{1.97}$_{\pm 0.21}$ & 0.0$_{\pm 0.0}$ & \better{128}$_{\pm 90}$ & \better{0.02}$_{\pm 0.00}$ & \textbf{45.3$_{\pm 25.1}$} & \better{1238}$_{\pm 198}$ & \better{0.82}$_{\pm 0.13}$ \\
flask                           & 9{,}314 & 482 & 0.0$_{\pm 0.0}$ & 7460$_{\pm 499}$ & 3.04$_{\pm 0.23}$ & 0.0$_{\pm 0.0}$ & \better{2786}$_{\pm 223}$ & \worse{5.30}$_{\pm 0.47}$ & 0.0$_{\pm 0.0}$ & \better{762}$_{\pm 190}$ & \better{0.60}$_{\pm 0.52}$ & 0.0$_{\pm 0.0}$ & \better{3173}$_{\pm 671}$ & \better{2.81}$_{\pm 0.81}$ \\
imapclient                      & 3{,}531 & 267 & 46.3$_{\pm 6.8}$ & 3482$_{\pm 530}$ & 1.13$_{\pm 0.04}$ & 49.0$_{\pm 2.2}$ & \better{1330}$_{\pm 548}$ & \worse{1.78}$_{\pm 0.18}$ & 3.3$_{\pm 4.7}$ & \better{261}$_{\pm 100}$ & \better{0.20}$_{\pm 0.12}$ & \textbf{63.3$_{\pm 2.0}$} & \better{1104}$_{\pm 345}$ & \better{0.54}$_{\pm 0.08}$ \\
parsel                          & 1{,}128 & 250 & 30.3$_{\pm 9.1}$ & 1503$_{\pm 75}$ & 0.55$_{\pm 0.12}$ & 14.3$_{\pm 10.5}$ & \better{1207}$_{\pm 566}$ & \worse{0.56}$_{\pm 0.22}$ & 0.0$_{\pm 0.0}$ & \better{444}$_{\pm 366}$ & \better{0.05}$_{\pm 0.01}$ & \textbf{56.7$_{\pm 8.2}$} & \better{1034}$_{\pm 73}$ & \better{0.34}$_{\pm 0.01}$ \\
portalocker                     & 1{,}958 & 71 & 0.0$_{\pm 0.0}$ & 1977$_{\pm 34}$ & 0.71$_{\pm 0.08}$ & 0.0$_{\pm 0.0}$ & \better{1055}$_{\pm 353}$ & \worse{0.86}$_{\pm 0.17}$ & 0.0$_{\pm 0.0}$ & \better{1485}$_{\pm 1134}$ & \better{0.66}$_{\pm 0.45}$ & 0.0$_{\pm 0.0}$ & \better{897}$_{\pm 74}$ & \better{0.35}$_{\pm 0.02}$ \\
pyjwt                           & 2{,}690 & 294 & 79.0$_{\pm 47.2}$ & 2222$_{\pm 275}$ & 0.79$_{\pm 0.06}$ & 126.0$_{\pm 15.1}$ & \better{1215}$_{\pm 203}$ & \worse{1.48}$_{\pm 0.11}$ & 29.0$_{\pm 41.0}$ & \better{156}$_{\pm 119}$ & \better{0.21}$_{\pm 0.25}$ & \textbf{140.0$_{\pm 9.1}$} & \better{1310}$_{\pm 95}$ & \better{0.50}$_{\pm 0.06}$ \\
python-hl7                      & 2{,}434 & 100 & 16.7$_{\pm 3.7}$ & 2534$_{\pm 586}$ & 0.78$_{\pm 0.25}$ & 33.3$_{\pm 8.0}$ & \better{2231}$_{\pm 28}$ & \worse{1.62}$_{\pm 0.03}$ & 20.3$_{\pm 14.7}$ & \better{1046}$_{\pm 827}$ & \worse{0.81}$_{\pm 0.56}$ & \textbf{42.3$_{\pm 6.6}$} & \better{1557}$_{\pm 249}$ & \better{0.58}$_{\pm 0.13}$ \\
rsa                             & 2{,}949 & 100 & 35.7$_{\pm 1.2}$ & 2640$_{\pm 183}$ & 0.93$_{\pm 0.13}$ & 43.7$_{\pm 2.9}$ & \better{1144}$_{\pm 166}$ & \worse{1.37}$_{\pm 0.10}$ & 13.3$_{\pm 18.9}$ & \better{211}$_{\pm 142}$ & \better{0.17}$_{\pm 0.19}$ & \textbf{54.3$_{\pm 5.9}$} & \better{1331}$_{\pm 195}$ & \better{0.62}$_{\pm 0.06}$ \\
simpy                           & 2{,}184 & 149 & 10.3$_{\pm 7.5}$ & 3147$_{\pm 267}$ & 1.42$_{\pm 0.17}$ & 39.3$_{\pm 3.4}$ & \worse{3472}$_{\pm 469}$ & \worse{1.89}$_{\pm 0.17}$ & 24.0$_{\pm 33.9}$ & \better{1081}$_{\pm 721}$ & \better{0.69}$_{\pm 0.48}$ & \textbf{58.0$_{\pm 7.5}$} & \better{1710}$_{\pm 269}$ & \better{0.68}$_{\pm 0.09}$ \\
tinydb                          & 2{,}170 & 204 & 110.0$_{\pm 18.6}$ & 2067$_{\pm 290}$ & 0.60$_{\pm 0.08}$ & 61.7$_{\pm 43.6}$ & \better{696}$_{\pm 12}$ & \worse{0.85}$_{\pm 0.01}$ & 67.3$_{\pm 95.2}$ & \better{301}$_{\pm 80}$ & \better{0.23}$_{\pm 0.14}$ & \textbf{125.7$_{\pm 21.4}$} & \better{656}$_{\pm 123}$ & \better{0.30}$_{\pm 0.03}$ \\
trailscraper                    & 890 & 93 & 14.7$_{\pm 2.5}$ & 2695$_{\pm 263}$ & 0.86$_{\pm 0.12}$ & 17.7$_{\pm 3.4}$ & \better{1562}$_{\pm 122}$ & \worse{1.34}$_{\pm 0.10}$ & 7.3$_{\pm 3.1}$ & \better{297}$_{\pm 139}$ & \better{0.30}$_{\pm 0.10}$ & \textbf{31.7$_{\pm 3.8}$} & \better{1237}$_{\pm 318}$ & \better{0.55}$_{\pm 0.06}$ \\
voluptuous                      & 3{,}100 & 161 & 38.7$_{\pm 4.0}$ & 2624$_{\pm 378}$ & 1.09$_{\pm 0.13}$ & 44.0$_{\pm 1.4}$ & \better{2400}$_{\pm 409}$ & \worse{1.62}$_{\pm 0.19}$ & 15.3$_{\pm 21.7}$ & \better{448}$_{\pm 338}$ & \better{0.70}$_{\pm 0.91}$ & \textbf{50.3$_{\pm 1.9}$} & \better{1566}$_{\pm 54}$ & \better{0.62}$_{\pm 0.05}$ \\
xmnlp                           & 1{,}504 & 23 & 0.7$_{\pm 0.5}$ & 3536$_{\pm 964}$ & 1.12$_{\pm 0.37}$ & 0.0$_{\pm 0.0}$ & \better{2456}$_{\pm 282}$ & \worse{3.37}$_{\pm 0.47}$ & 0.0$_{\pm 0.0}$ & \better{2043}$_{\pm 1666}$ & \better{0.62}$_{\pm 0.58}$ & \textbf{2.7$_{\pm 0.9}$} & \better{1250}$_{\pm 342}$ & \better{0.96}$_{\pm 0.14}$ \\
zxcvbn                          & 1{,}386 & 31 & 13.0$_{\pm 1.4}$ & 1479$_{\pm 32}$ & 0.66$_{\pm 0.13}$ & 14.0$_{\pm 0.8}$ & \better{854}$_{\pm 60}$ & \worse{0.87}$_{\pm 0.13}$ & \textbf{16.3$_{\pm 9.7}$} & \better{306}$_{\pm 113}$ & \better{0.14}$_{\pm 0.02}$ & \textbf{16.3$_{\pm 0.5}$} & \better{823}$_{\pm 157}$ & \better{0.50}$_{\pm 0.10}$ \\
\midrule
Avg                             & -- & -- & 20.1\% & 2756 ($1\times$) & 1.03 ($1\times$) & 23.3\% & 1771 (\better{$0.64\times$}) & 1.65 (\worse{$1.60\times$}) & 16.3\% & \textbf{680 (\better{$0.25\times$})} & \textbf{0.38 (\better{$0.37\times$})} & \textbf{34.1\%} & 1315 (\better{$0.48\times$}) & 0.67 (\better{$0.65\times$}) \\
\bottomrule
\end{tabular}}
\end{table}

\paragraph{Efficiency trade-offs.} Across both benchmarks, a consistent pattern emerges: File-based Parallel trades cost for speed without improving quality, Claude Code w/ Agent Teams trades quality for speed, and \method improves all three metrics simultaneously. The key mechanism is the cohesion-based graph partition, which groups tightly coupled files to minimize inter-agent communication, combined with the event-driven shared task list that resolves inter-group dependencies without global synchronization. This design enables agents to proceed as soon as upstream files are ready, achieving the latency benefits of parallelism while preserving the interface consistency that correctness requires.

\section{Discussion}
\label{sec:discussion}

\paragraph{Why graph partition gains scale with project complexity.}
The advantage of cohesion-aware partitioning grows with the density of inter-file coupling. In simple projects (e.g., \texttt{Hybrid\_Images} with 19 tests across loosely connected files), most methods can produce consistent interfaces because files share few type contracts. As project size increases, however, cross-file dependencies grow super-linearly: a file that imports symbols from $k$ other files creates $k$ potential points of interface mismatch when those files are generated independently. This trend is supported by our analysis: across the 16 CodeProjectEval projects with non-zero pass rates, edge density ($|E|/|V|$; Table~\ref{tab:partition_stats}) correlates positively with absolute test-pass improvement over Sequential (Pearson $r{=}0.65$, Spearman $\rho{=}0.60$, both $p{<}0.05$). Na\"ive parallelism exposes these dependency edges to concurrent, uncoordinated generation, so conflicts become increasingly likely as coupling rises. In contrast, \method's cohesion-based partition places strongly coupled files within the same agent, internalizing the majority of dependency edges while resolving the remainder through the shared task list's topological ordering.

\paragraph{Latency--correctness trade-off.}
Our results reveal a fundamental tension between minimizing wall-clock latency and maximizing code correctness. Claude Code w/ Agent Teams achieves the lowest latency among baselines on CodeProjectEval by aggressively distributing work across independent agents, but this comes at the cost of the lowest pass rate---falling below even the sequential baseline on CodeProjectEval (16.3\% vs.\ 20.1\%). The sequential baseline, conversely, maintains interface consistency but cannot exploit any parallelism. \method navigates this trade-off through event-driven scheduling: rather than launching all tasks simultaneously or executing them serially, it releases each file for generation as soon as its upstream dependencies are satisfied. This yields latency reductions of 45--52\% over Sequential while simultaneously improving pass rate, because agents begin work with access to the actual interfaces produced by upstream files rather than speculative placeholders.

\paragraph{Limitations.}
Two scope boundaries constrain the current system. First, when a project exhibits near-complete coupling (i.e., almost every file depends on almost every other), the partition algorithm produces a single group, and execution degrades to sequential. In such cases \method offers no latency advantage over the sequential baseline, though it retains the benefit of structured dependency resolution. Second, our evaluation is limited to Python repositories; whether the observed gains transfer to statically typed languages with richer compiler feedback (e.g., Rust, Java) remains an open question for future work.
\section{Related Works}
\subsection{Repository-level Coding}

LLMs handle small, self-contained programming tasks well \cite{jain2024livecodebench} but degrade sharply on real-world repository-scale software engineering, where modifications must be coordinated across many interdependent files \cite{jimenez2023swebench, liu2023repobench}. Recent agent systems mitigate this with structured agent-computer interfaces and staged pipelines: SWE-agent \cite{yang2024sweagent} and OpenHands \cite{wang2024openhands} expose controlled tool environments for iterative patch refinement, while Agentless \cite{xia2024agentless} decomposes bug resolution into localization, repair, and validation stages with explicit checkpoints. A complementary line tackles context management through retrieval-augmented generation \cite{wu2024repoformer}, though selecting informative context over a large codebase in long-horizon remains an open problem \cite{yang2024sweagent, wang2024openhands}.

\subsection{Multi-agent Systems Orchestration}

LLM-based multi-agent orchestration has largely prioritized task quality through role specialization and structured communication. Organizational-style frameworks assign agents to specialized roles connected by predefined workflows \cite{hong2024metagpt, qian2024chatdev}; conversational frameworks generalize coordination to free-form dialogue between agents and arbitrary topologies \cite{li2023camel, wu2023autogen}; and graph-structured approaches encode agent interactions as optimizable graphs or DAG-based collaboration networks to improve scaling \cite{zhuge2024gptswarm, qian2025macnet}. Across these designs, orchestration is performance-oriented and relies on sequential handoffs or runtime communication protocols, so wall-clock latency scales with task complexity rather than with available agents. Recent diagnostic studies trace this to context mismanagement, shallow communication, and a spatial-semantic coordination asymmetry in which agents agree on \emph{where} to edit but fail to coordinate \emph{what} to implement \cite{cemri2025why, malfa2025large, khatua2026cooperbench}, indicating that the bottleneck is structural rather than communicative. \method addresses this by reframing orchestration as a graph partitioning problem, assigning each agent a structurally cohesive group of files and eliminating most cross-agent dependencies before execution begins.

\subsection{Efficient Planning}

Efficient decomposition and scheduling of dependent tasks has long been studied in classical settings, from industrial workflow allocation \cite{kallrath2002planning} to heterogeneous distributed computing \cite{topcuoglu2002performance}, where DAG-based execution and critical-path methods are standard tools for minimizing wall-clock latency under dependency constraints. Recent LLM-based systems adopt similar structures: graph-enhanced asynchronous plan decomposition \cite{lin2024graph} and temporal constraint-based planning \cite{ding-etal-2025-tcp} for general reasoning, CodePlan \cite{bairi2024codeplan} for single-agent repository editing via DAG-ordered file edits, and LLMCompiler \cite{kim2024llmcompiler} for parallel function-call execution within a single agent. \method extends these paradigms to the multi-agent setup where the dependency graph simultaneously drives task partitioning across agents and the execution order, coupling allocation and scheduling into a single optimization problem.

\section{Conclusion}
We identify task partitioning as a key bottleneck in multi-agent LLM coordination and cast it as a graph optimization problem capturing the communication-to-computation trade-off. Our method, \method, achieves up to $2.10\times$ wall-clock speedup and improved pass rate on \devbench and CodeProjectEval, with the largest gains on the most dependency-dense projects. This suggests cohesion-based orchestration as a broader design principle for multi-agent LLM systems.



\bibliographystyle{plain}
\bibliography{reference}

\clearpage
\appendix

\section{Quality Control of Repository Interface Blueprint}
\label{sec:rib_validation}

The correctness of the Repository Interface Blueprint (RIB) is critical, as all subsequent steps---coupling weight estimation, graph partitioning, and parallel scheduling---depend on it. We employ an LLM-based judge to score the generated architecture on a 1--10 scale across three dimensions: (i) completeness of interface definitions, (ii) correctness of dependency edges, and (iii) consistency of type annotations. If the average score falls below a threshold (default 7), the RIB is refined by feeding the judge's critique back to the generation model, and re-judged. This loop iterates for up to three rounds.

\section{Implementation Details}
\label{app:impl}

Table~\ref{tab:hyperparams} lists the hyperparameters used throughout all experiments. All values are fixed across repositories and benchmarks; no per-project tuning is performed.

\begin{table}[t]
\centering
\caption{Hyperparameters of \method. All values are held constant across the 28 evaluated repositories.}
\label{tab:hyperparams}
\small
\begin{tabular}{llc}
\toprule
Component & Parameter & Value \\
\midrule
\multirow{4}{*}{Edge weight (Eq.~\ref{eq:cosweight})}
  & Defined-symbol weight in $\mathbf{s}_i$ & 2 \\
  & Referenced-type weight in $\mathbf{s}_i$ & 1 \\
  & Cosine scale $\gamma$ & 10 \\
  & Default weight (no shared symbols) & 0.1 \\
\midrule
\multirow{1}{*}{Community detection (Sec.~\ref{sec:partition})}
  & Markov time $t$ & 1.0 \\
\midrule
\multirow{1}{*}{Hub isolation (Sec.~\ref{sec:partition})}
  & Role threshold (fan-in/fan-out ratio) & 0.4 \\
\midrule
\multirow{1}{*}{RIB generation (Sec.~\ref{sec:graph})}
  & Maximum refinement rounds & 3 \\
\midrule
\multirow{1}{*}{Test-driven repair (Sec.~\ref{sec:execution})}
  & Maximum repair iterations & 10 \\
\bottomrule
\end{tabular}
\end{table}

\paragraph{Edge weight construction.} Each file's symbol vector $\mathbf{s}_i$ assigns weight~2 to symbols the file \emph{defines} (class names, top-level functions, global constants) and weight~1 to types it \emph{references} in parameter annotations, after filtering builtins. The cosine similarity is scaled by $\gamma{=}10$ to place edge weights in the same numerical range as the trade-off coefficient $\alpha$ in Eq.~\eqref{eq:objective}. Edges between files that share no symbols receive a floor weight of $0.1$ to preserve graph connectivity.

\paragraph{Role classification.} A file is classified as an \emph{in-hub} if its in-degree ratio $|\{v_j : (v_j, v_i) \in E\}|\,/\,(|V|-1)$ exceeds $0.4$, indicating it is widely depended upon; and as an \emph{out-hub} (integration file) if its out-degree ratio $|\{v_j : (v_i, v_j) \in E\}|\,/\,(|V|-1)$ exceeds the same threshold, indicating it aggregates many upstream modules. The threshold is intentionally conservative: raising it merges more utility files into community clusters, increasing intra-group serialization; lowering it isolates too many files into singletons, fragmenting cohesive modules.

\section{Per-Repository Partition Statistics}
\label{app:partition_stats}

Table~\ref{tab:partition_stats} characterizes the dependency graph $G{=}(V,E)$ and the partition produced by Algorithm~\ref{alg:partition} for each repository. CodeProjectEval projects exhibit markedly higher edge density ($|E|/|V| \approx 1.5$) than \devbench ($|E|/|V| \approx 0.3$), explaining why cohesion-aware partitioning yields larger gains on CodeProjectEval (Section~\ref{sec:results}).

\paragraph{Note on file and LOC counts.} The \emph{\#Files} and \emph{\#LOC} columns report only source files that agents are tasked to generate. A small number of \texttt{.py} files in the original benchmarks are auto-generated artifacts (e.g., Django database migrations in \texttt{drf-simplejwt}, static data tables in \texttt{zxcvbn}) or packaging scripts (\texttt{setup.py}) that do not contain meaningful implementation logic; these are pre-populated in the agent workspace and excluded from the generation task. Consequently, the file counts for a few repositories differ slightly from those reported in the original benchmark papers.

\begin{table}[h]
\centering
\caption{Dependency graph and partition statistics. \emph{\#Files} and \emph{\#LOC} count only source files that agents must generate; auto-generated data files (e.g., Django migrations, static lookup tables) are pre-populated in the workspace and excluded from the generation task. \emph{\#Edges}: directed dependency edges $|E|$; \emph{Density}: edge density $|E|/|V|$; \emph{\#Hubs}: structural hubs isolated per Section~\ref{sec:partition}; \emph{\#Groups}: total partitions $K$; \emph{MaxGrp}: largest cohesive cluster size; \emph{\#Sing}: singleton partitions (hubs + files lifted for exploiting latent parallelism). }
\label{tab:partition_stats}
\small
\setlength{\tabcolsep}{4pt}
\begin{tabular}{l rr rr rrrr}
\toprule
Task & \#Files & \#LOC & \#Edges & Density & \#Hubs & \#Groups & MaxGrp & \#Sing \\
\midrule
\multicolumn{9}{l}{\textit{\devbench}} \\
\midrule
ArXiv\_digest   & 2 & 197 & 0 & 0.00 & 0 & 1 & 1 & 1 \\
chakin          & 2 & 62 & 1 & 0.33 & 0 & 2 & 2 & 1 \\
geotext         & 2 & 134 & 0 & 0.00 & 0 & 1 & 2 & 0 \\
hone            & 6 & 271 & 1 & 0.14 & 0 & 4 & 4 & 3 \\
Hybrid\_Images  & 1 & 144 & 0 & 0.00 & 0 & 1 & 1 & 1 \\
lice            & 2 & 374 & 1 & 0.50 & 0 & 1 & 2 & 0 \\
pso             & 3 & 166 & 0 & 0.00 & 0 & 2 & 2 & 1 \\
readtime        & 5 & 288 & 2 & 0.40 & 0 & 3 & 3 & 2 \\
stocktrends     & 2 & 383 & 0 & 0.00 & 0 & 8 & 2 & 7 \\
TextCNN         & 6 & 400 & 4 & 0.80 & 0 & 5 & 1 & 5 \\
\midrule
\multicolumn{9}{l}{\textit{CodeProjectEval}} \\
\midrule
bplustree       & 8 & 1{,}509 & 15 & 1.88 & 1 & 5 & 3 & 3 \\
cookiecutter    & 19 & 2{,}811 & 36 & 2.00 & 2 & 10 & 4 & 6 \\
csvs-to-sqlite  & 3 & 816 & 1 & 0.33 & 1 & 2 & 2 & 1 \\
deprecated      & 3 & 597 & 2 & 0.67 & 1 & 2 & 2 & 1 \\
drf-simplejwt   & 18 & 1{,}712 & 29 & 1.61 & 2 & 11 & 5 & 7 \\
flask           & 24 & 9{,}308 & 64 & 2.67 & 1 & 8 & 8 & 3 \\
imapclient      & 17 & 3{,}531 & 25 & 1.47 & 0 & 10 & 5 & 6 \\
parsel          & 5 & 1{,}128 & 5 & 1.00 & 0 & 4 & 2 & 3 \\
portalocker     & 9 & 1{,}958 & 16 & 1.78 & 2 & 7 & 3 & 6 \\
pyjwt           & 12 & 2{,}690 & 27 & 2.25 & 1 & 6 & 4 & 3 \\
python-hl7      & 11 & 2{,}434 & 16 & 1.45 & 0 & 4 & 5 & 2 \\
rsa             & 14 & 2{,}949 & 25 & 1.79 & 1 & 10 & 3 & 7 \\
simpy           & 12 & 2{,}184 & 20 & 2.00 & 1 & 7 & 4 & 6 \\
tinydb          & 10 & 2{,}170 & 11 & 1.10 & 0 & 6 & 4 & 4 \\
trailscraper    & 13 & 890 & 16 & 1.23 & 0 & 8 & 2 & 3 \\
voluptuous      & 6 & 3{,}100 & 12 & 2.00 & 2 & 4 & 3 & 3 \\
xmnlp           & 24 & 1{,}504 & 23 & 0.96 & 0 & 9 & 5 & 0 \\
zxcvbn          & 6 & 1{,}386 & 6 & 1.00 & 1 & 5 & 2 & 4 \\
\bottomrule
\end{tabular}
\end{table}

\section{Clarification on Benchmark Naming}
\label{app:naming}

This paper evaluates on two benchmarks: \devbench~\cite{deveval} and CodeProjectEval~\cite{zhao2025realisticprojectlevelcodegeneration}.  Note that \devbench was formerly named as ``DevBench'' in its earlier arXiv version. Readers should be aware that other benchmarks~\cite{li2024deveval, devbench2, devbench3} share the same names.

\section{Repository Interface Blueprint Example}
\label{app:rib_example}

Figure~\ref{fig:rib_example} shows a condensed excerpt of the Repository Interface Blueprint (RIB) produced for the \texttt{zxcvbn} project (6 files, 6 dependency edges). The RIB specifies, for each file: (i) the files it depends on, (ii) a natural-language description of its role, and (iii) the function signatures it defines. This structured representation provides the inputs to graph construction (Section~\ref{sec:graph}): dependencies yield the edge set $E$, and shared symbols across function signatures yield the coupling weights $c_{ij}$ via Eq.~\eqref{eq:cosweight}.

\begin{figure}[h]
\begin{jsonbox}[RIB excerpt --- \texttt{zxcvbn} (6 files, 6 dependency edges)]
[{
  "name": "zxcvbn",
  "files": [
    {
      "path": "zxcvbn/__init__.py",
      "dependencies": [
        "zxcvbn/feedback.py", "zxcvbn/matching.py",
        "zxcvbn/scoring.py", "zxcvbn/time_estimates.py"
      ],
      "description": "Package initializer exposing the public API.",
      "functions": [{
        "name": "zxcvbn",
        "parameters": [
          {"name": "password", "type": "str"},
          {"name": "user_inputs", "type": "list[str] or None"},
          {"name": "max_length", "type": "int", "default": "72"}
        ]
      }]
    },
    {
      "path": "zxcvbn/scoring.py",
      "dependencies": [],
      "description": "Guess estimation and match sequence optimization.",
      "functions": [
        {"name": "most_guessable_match_sequence", "parameters": ["..."]},
        {"name": "estimate_guesses", "parameters": ["..."]}
      ]
    },
    {
      "path": "zxcvbn/matching.py",
      "dependencies": ["zxcvbn/scoring.py"],
      "description": "Pattern detection (dictionary, spatial, sequence, etc.).",
      "functions": [
        {"name": "omnimatch", "parameters": ["..."]},
        {"name": "dictionary_match", "parameters": ["..."]}
      ]
    }
  ]
}]
\end{jsonbox}
\caption{Condensed RIB for the \texttt{zxcvbn} project. Three of six files shown; remaining files (\texttt{feedback.py}, \texttt{time\_estimates.py}, \texttt{\_\_main\_\_.py}) omitted for space.}
\label{fig:rib_example}
\end{figure}


\clearpage

\end{document}